\documentclass[10pt,conference,a4paper]{IEEEtran}
\usepackage{times,amsmath,epsfig}
\usepackage{epstopdf}
\usepackage{booktabs}
\usepackage{amsmath}
\newcommand{\bftab}{\fontseries{b}\selectfont}
\title{Long-Term Video Interpolation with Bidirectional Predictive Network}
\author{%
{Xiongtao Chen,Wenmin Wang,Jinzhuo Wang,Weimian Li,Baoyang Chen}
\vspace{1.6mm}\\
\fontsize{10}{10}\selectfont\itshape
\ Peking University\\ %  Removed for anonymous submission
\fontsize{9}{9}\selectfont\ttfamily\upshape
\vspace{1.2mm}\\
\fontsize{10}{10}\selectfont\rmfamily\itshape
\fontsize{9}{9}\selectfont\ttfamily\upshape
\,
}
\begin{document}
\maketitle

 %INCLUDES COPYRIGHT NOTICE: one of three copyright notice should be included. Uncomment the appropriate line below, according to the authors affiliation:
%\begin{figure}[b]
%\parbox{\hsize}{\em
%%information about the event:
%%IEEE VCIP'14, Dec. 7 - Dec. 10, 2014, Valletta, Malta.
%
%%copyright notice: one of three copyright notices below should be included. Uncomment the appropriate line, according to the authors affiliation:
%%000-0-0000-0000-0/00/\$31.00 \ \copyright 2014 IEEE.
%%U.S. Government work not protected by U.S. copyright.
%%???-?-????-????-?/10/\$??.?? \copyright 2014 Crown.
%}\end{figure}

\begin{abstract}
This paper considers the challenging task of long-term video interpolation. Unlike most existing methods that only generate few intermediate frames between existing adjacent ones, we attempt to speculate or imagine the procedure of an episode and further generate multiple frames between two non-consecutive frames in videos. In this paper, we present a novel deep architecture called bidirectional predictive network (BiPN) that predicts intermediate frames from two opposite directions. The bidirectional architecture allows the model to learn scene transformation with time as well as generate longer video sequences. Besides, our model can be extended to predict multiple possible procedures by sampling different noise vectors. A joint loss composed of clues in image and feature spaces and adversarial loss is designed to train our model. We demonstrate the advantages of BiPN on two benchmarks Moving 2D Shapes and UCF101 and report competitive results to recent approaches.
%The typical methods for video interpolation and extrapolation often utilize optical flow algorithms that only work when optical flow is accurately estimated. CNNs have recently been applied to video prediction successfully but can only produce a few frames in very short-term.
%Predictive models have recently achieved promising results in video prediction but can only produce a few frames in very short-term.
% generative adversarial network with a bidirectional encoder-decoder architecture that predict intermediate frames from two opposite direction. The bidirectional architecture allows us to generate longer video sequences while adversarial training manner contribute to realistic results. Besides, our model can be extended to predict multiple possible procedures by sampling noise vectors. We design a joint loss of reconstruction error, adversarial error and feature-space loss to train our model. We demonstrate the advantages of BiPN on two benchmarks Moving 2D Shapes and UCF-101 and report competitive results to the state-of-the-art.
%%We present an unsupervised method to address the problem of predicting long-term in-between existing frames.
\\[1\baselineskip]
\end{abstract}

%% NOTE keywords are not used for conference papers so do not populate them
%\begin{keywords}
%
%\end{keywords}
%%

\section{Introduction}
\begin{figure}[t]
  \centering
  %\fbox{\rule[-.5cm]{0cm}{3cm} \rule[-.5cm]{8cm}{0cm}}
  \includegraphics[width=8.5cm]{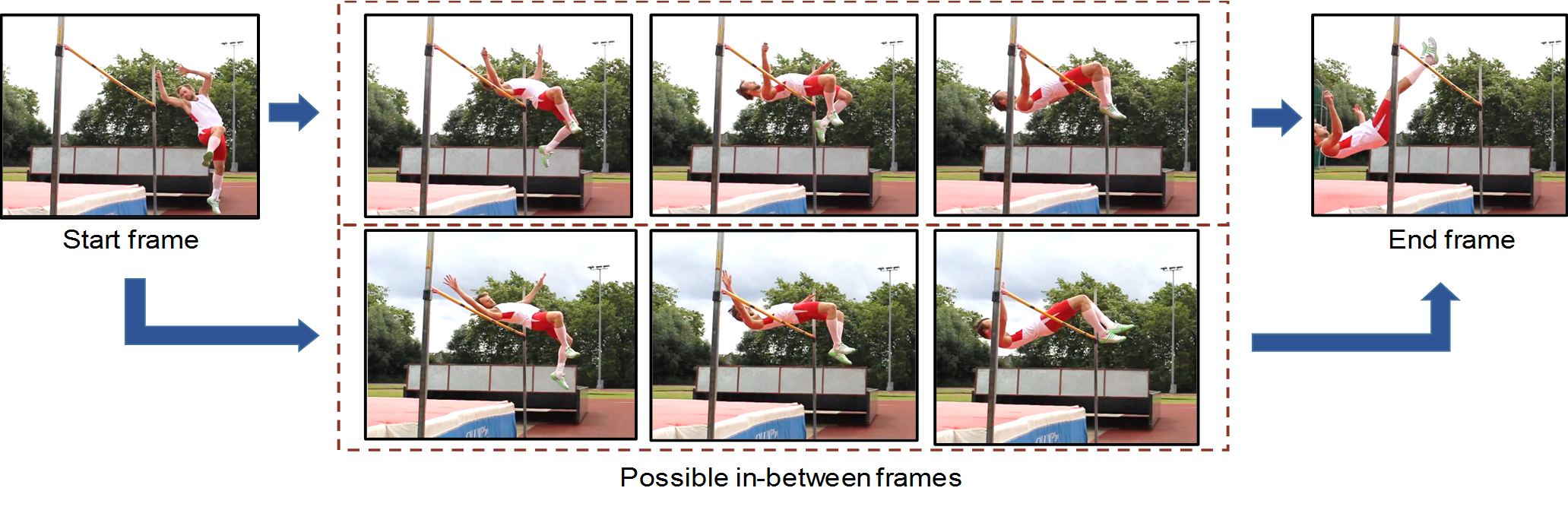}
  \caption{Task of video interpolation. Given a start frame and end frame, humans can speculate or imagine several possible in-between frames. Here shows two possible ways how the high jumper gets over the bar.}\label{fig:fig-ex1}
\end{figure}

% 对于视觉世界，人类具有推测想象的能力。 我们探索计算机视觉算法是否可以做同样的事情。
Video understanding has been one of the most important tasks in computer vision. Compared to still images, the temporal component of videos provides richer descriptions of visual world, which offers possibilities to predict unknown situations.  By observing two nonadjacent frames in a natural video, humans have an uncanny ability to speculate what happens in the intermediate frames. As shown in Fig. \ref{fig:fig-ex1}, in spite of the loss of in-between frames, we can still easily imagine how the high jumper jumps up, clears the bar and falls down. In this paper, we explore whether machine learning algorithms can be endowed with this ability to predict events and further perform long-term interpolation in videos. Long-term video interpolation techniques can not only be applied to frame rate conversion in video or film production, but also provide potentials for missing frames recovery, new scene generation and anomaly detection in surveillance videos.

% 传统的插值算法，最近的视屏预测方法。 我们结合预测方法用于long―term interpolation
Most traditional frame interpolation methods utilize optical flow algorithms to estimate dense motion between two consecutive frames and then interpolate along optical flow vectors \cite{baker2011database} \cite{zhang2004survey} \cite{revaud2015epicflow}. However, these methods require accurate estimation of dense correspondence which is challenging for large and fast motions. \cite{niklaus2017video} employs a deep fully CNN to combine estimation motion and pixel synthesis into a single process. Deep voxel flow (DVF) method proposed by \cite{liu2017video} flows pixels from existing frames to produce new frames. Recently, predictive models have received increasing attention for video prediction.  Advanced works have developed kinds of generative models to anticipate future frames and report promising results \cite{mathieu2015deep} \cite{vondrick2016generating}.

% 我们的主要方法
In this paper, we aim to tackle the more difficult task of long-term video interpolation, where the missing frames can not receive enough ``hints'' from nearby frames. We borrow the idea from the predictive models and formulate the process of interpolation as one of bidirectional prediction. Given two non-consecutive frames, we train a convolutional encoder-decoder network to regress to the missing intermediate frames from two opposite directions. We refer to our model as \textbf{bidirectional predictive network (BiPN)}, as it consists of a bidirectional encoder-decoder that predicts the future forward from start frame and ``predicts'' the past backward from end frame at the same time. The bidirectional architecture comes from two critical considerations: (1) Intermediate frames generation requires understanding appearance and motion signals from both start frame and end frame to learn scene transform with time. (2) High-quality long-term predictions are still difficult to achieve since the noise amplifies quickly through time and the prediction degrades dramatically as argued in \cite{villegas2017learning}. Predicting in two direction is a reasonable idea to synthesize roughly double decent intermediate frames.
% multi-scale architecture.

% 多模态问题
Our model can understand the input frames and produce a plausible hypothesis for the missing intermediate frames. However, there may exist multiple ways to recover the missing frames as humans can usually imagine multiple paths to get to the final state. As we can see in Fig. \ref{fig:fig-ex1}, a high jumper may get over the bar in different poses. Therefore, this task is inherently multi-modal and much difficult. We make attempts to tackle this multi-modal problem by extending the BiPN with noise vector input. See Section II.C for details.
 %Similar to the architecture of conditional GAN \cite{mirza2014conditional}, our network can be extended to fulfill this multi-modal requirement by adding a noise vector. The noise vector is sampled from Gaussian distribution and different noises can result in different outputs for a specific input.

% 训练，joint loss function

To produce such frames with more accurate appearance and realistic looking, we combine a loss in image space and feature space with an adversarial loss \cite{goodfellow2014generative} to train our network. Similar efforts have demonstrated effectivity for generative models such as inpainting \cite{pathak2016context} and new frames synthesis \cite{villegas2017learning}.

The main contributions of this paper are summarized as follows. First, we propose a deep bidirectional predictive network (BiPN) for the challenging task of long-term video interpolation. Second, we extend the BiPN to deal with the multi-modal problem aiming to mimic the diversity of  human imagination. Finally, we evaluate our model on a synthetic dataset Moving 2D Shapes and a natural video dataset UCF101 and report competitive results compared to recent approaches both in quantity and quality.

\section{Approach}
We now introduce BiPN, a convolutional encoder-decoder that can predict long-term intermediate frames from two non-consecutive frames. We first present the general architecture and a multi-scale version of BiPN for single procedure generation, then extend the model to tackle the multi-model issue and finally describe the details of training and implementation.

\subsection{Bidirectional Predictive Network (BiPN)}
\begin{figure}[t]
  \centering
  %\fbox{\rule[-.5cm]{0cm}{4cm} \rule[-.5cm]{8.5cm}{0cm}}
  \includegraphics[width=8.5cm]{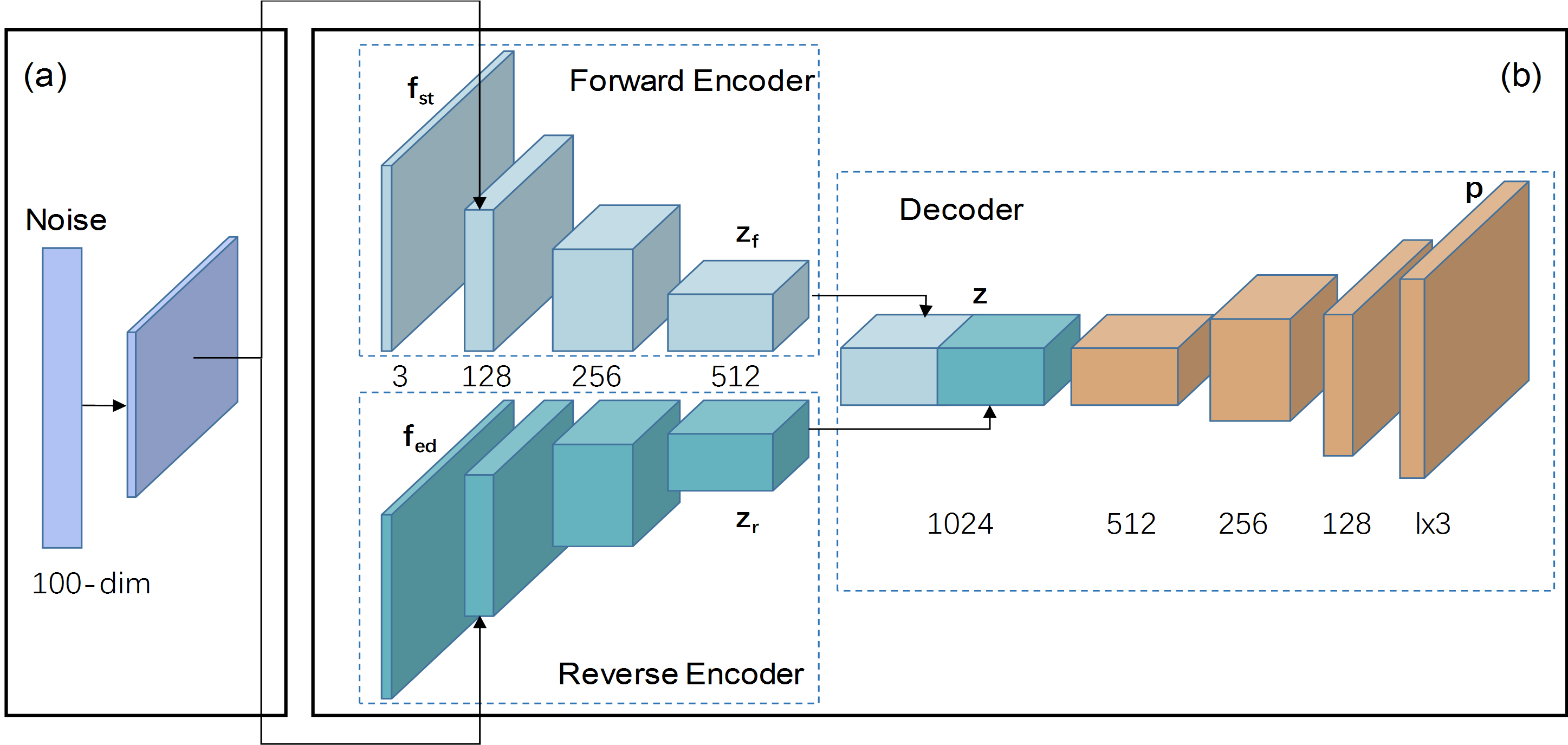}
  \caption{Bidirectional Predictive Network (BiPN). (a) Extended noise input to the network to produce multiple possible intermediate procedures. (b) General architecture of BiPN. The given started frame $f_{st}$ and end frame $f_{ed}$ are passed through the forward encoder and reverse encoder respectively to obtain features $z_f$ and $z_r$, which are connected to the decoder after concatenated. The decoder finally predicts $l$ in-between frames $p$.}\label{fig:fig-model}
\end{figure}

The general architecture of our BiPN is an encoder-decoder pipeline, including a bidirectional encoder and a single decoder. The bidirectional encoder encodes information from both start frame and end frame and produces a latent feature representation. The decoder takes this feature representation as input to predict multiple missing frames between the two given frames. Fig. \ref{fig:fig-model}(b) shows an overview of our architecture.

\textbf{Bidirectional Encoder.}
Intuitively, it is inappropriate to predict in-between sequences only utilizing the start frame or the end frame. Instead, predicting multiple intermediate frames requires to understand the appearance and motion signals from both the start and end frames. To this end, we design our encoder as a bidirectional structure, including a forward encoder and a reverse encoder. The forward encoder takes the start frame $f_{st}$ as input and extracts an abstract feature representation $z_f$ while the reverse encoder produces representation $z_r$ from the end frame $f_{ed}$. Notably, the channels of the end frame are reversed before fed into the reverse encoder. The forward and reverse encoders have the same network architecture, both consisting of several convolutional layers, each with a rectified linear unit (ReLU)  followed. The $z_f$ and $z_r$ are then concatenated into one latent representation $z$.
%activation function

\textbf{Decoder.}
The second half of our pipeline is a single decoder, which processes the latent feature representation $z$ from the above encoder to generate pixels of the in-between frames. The decoder is composed of a series of up-convolutional layers \cite{dosovitskiy2015learning} and ReLUs.
%A up-convolutional layer is a convolution that results in a higher resolution image, which can be understood as up-sampling followed by convolution as \cite{dosovitskiy2015learning} suggests.
Finally, the decoder outputs a feature map with the size of $l\times h \times w \times c$ as the prediction of the target in-between frames, where $l$ is the length of frames to be predicted, $h$, $w$ and $c$ are the height, width and the number of channels respectively for each frame ($c=3$ for RGB images).

The BiPN has the advantage to learn temporal dependencies from both start frame and end frame. The bidirectional architecture also comes from another critical consideration: most generative CNNs can only produce decent first few predictions and then the prediction degrades dramatically. Predicting from the start frame and the end frame bidirectionally can synthesize roughly double high quality in-between frames.

% synthesizing  sharp long-term future frames is difficult for . [These methods that directly hallucinate RGB pixel values of synthesized video frames can produce decent first few predictions but then the prediction degrades dramatically until all the video context lost, since the noise amplifies quickly through time until it overwhelms the signal \cite{}].

\subsection{Multi-scale Architecture}
Limited by the size of the kernels, convolutions only account for short-range dependencies. This limitation bring difficulties for generative CNNs to encode both large and small motions in a given scene. To tackle this problem, we enrich our BiPN into a multi-scale architecture, which has been adopted in many prediction works \cite{mathieu2015deep} \cite{liu2017video}  and proved to be effective.

\begin{figure}[t]
  \centering
  %\fbox{\rule[-.5cm]{0cm}{3cm} \rule[-.5cm]{8cm}{0cm}}
  \includegraphics[height=3cm]{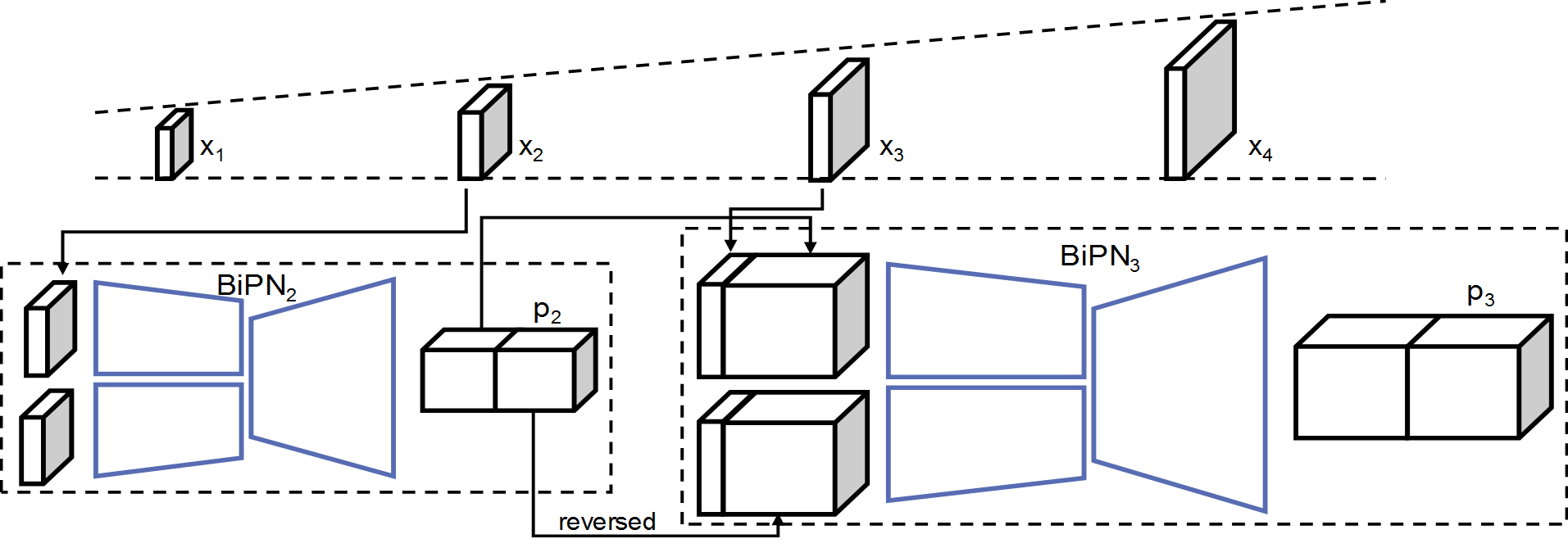}
  \caption{Multi-scale BiPN. Intermediate frames are obtained from a coarse to fine scale. Prediction from lower resolutions are split into two halves and next concatenated to start frame and end frame at higher resolutions as auxiliary inputs. We only show the details of scale $2$ and $3$ here.}\label{fig:fig-multiscale}
\end{figure}

We employ $4$ scales ($k=1,2,3,4$) and design $4$ BiPNs in practice. The multi-scale architecture is shown in Fig. \ref{fig:fig-multiscale}. Given an origin frame of size $s\times s$ for scale $k=4$, denoted as $x_4$, the frame is resized to $\frac{1}{2}s\times \frac{1}{2}s$, $\frac{1}{4}s\times \frac{1}{4}s$ and $\frac{1}{8}s\times \frac{1}{8}s$ images (defined as $x_3, x_2, x_1$) for $k=3,2,1$ respectively.
%the frame is resized to $\frac{1}{2}s\times \frac{1}{2}s$ image ($x_3$) for scale $k=3$, $\frac{1}{4}s\times \frac{1}{4}s$ image ($x_2$) for $k=2$ and $\frac{1}{8}s\times \frac{1}{8}s$ image ($x_1$) for $k=1$.
These images of different sizes are fed into different BiPNs to produce multiple scales of in-between frames $p_k$. It is worth noting that these BiPNs do not process images $x_{1,2,3,4}$ in a parallel but a sequential way. In more details, the first prediction $p_1$ produced by $\textnormal{BiPN}_1$ is upsampled to  $\frac{1}{4}s\times \frac{1}{4}s$ and then concatenated to $x_2$ as input of $\textnormal{BiPN}_2$. In the same way, $p_2$ and $p_3$ are used as auxiliary inputs to $\textnormal{BiPN}_3$ and $\textnormal{BiPN}_4$. The output $p_4$ of $\textnormal{BiPN}_4$ is chosen as the final prediction.
%of the in-between frames.

\subsection{Multi-modal Procedures}
Given an initial state and a final state, the BiPN presented above can only predict one possible procedure for a specific scene. However, in most cases, people tend to imagine multiple potential pathways or situations to get to the target state as shown in Fig. \ref{fig:fig-ex1}. To mimic this human behavior, we extend our BiPN to explore the space of possible actions by adding a random Gaussian noise to the encoder. Passing the start and end frames as input and sampling from the noise variable allow the model to predict multiple possible sets of in-between frames according to multiple different input noises. The extended part of the architecture is shown in Fig. \ref{fig:fig-model}. To the best of our knowledge, we are the first to address the multi-modal problem in long-term video interpolation.

\subsection{Training}
We train BiPN by regressing to the ground truth in-between frames. We clip each video into small sequences to construct training and test sets. Given a training sequence $X=\{x_1, x_2, ..., x_T\}$, we use $x_1$ and $x_T$ as start frame and end frame. The remaining frames are used as ground truth.

To synthesize high-quality and realistic frames, we optimize our network to minimize distance between the predicted frames and target frames in both image space and feature space, as well as minimize the adversarial loss \cite{goodfellow2014generative}. Thus the joint loss function of our network can be defined as:
\begin{equation}
      L = L_{rec} + L_{feat} + L_{adv},
\end{equation}
where $L_{rec}$ is the $L_2$ reconstruction loss responsible for producing a rough outline of the predicted frames. However, $L_2$ loss often results in blurry averaged images. $L_{feat}$ is the loss in feature space aiming to predict frames with more accurate appearance. We use the feature maps from the last convolutional layer of AlexNet \cite{krizhevsky2012imagenet}. $L_{adv}$ is the adversarial loss, which is introduced by GAN \cite{goodfellow2014generative} and has been adopted by many recent generative models to produce realistic images. Since our BiPN produces a small video sequence instead of a single image each time, we need to design a discriminator $D$ that tries to distinguish real and fake videos. We utilize multiple spatio-temporal convolutional layers to construct $D$. The $L_{rec}$, $L_{feat}$ and $L_{adv}$ can be expressed as following:
\begin{equation}
     L_{rec} = \sum\nolimits_{k=1}^{N_{scales}} \left\| p_k-\hat{p_k} \right\|_2^2,
\end{equation}

\begin{equation}
     L_{feat} = \sum\nolimits_{k=1}^{N_{scales}} \left\| Alex(p_k)- Alex(\hat{p_k}) \right\|_2^2,
\end{equation}
\begin{equation}
     %L_{adv} = \sum\nolimits_{k=1}^{N_{scales}}-\log{{D_k}(p_k, \hat{p_k})},
     L_{adv} = \sum\nolimits_{k=1}^{N_{scales}}L_{bce}(D_k(\hat{p_k}), 1)
\end{equation}
where $p_k$ and $\hat{p_k}$ are ground truth frames and our prediction respectively at scale $k$. $Alex(\cdot)$ indicates extracted features from all images in the sequence using AlexNet. $L_{bce}$ is the binary cross-entropy loss. During the multi-modal training, $L_{rec}$ is not used since it will restrict the diversity of outputs.

%\begin{equation}
%    \textnormal{L}_{loc(p_t,g_c)=\textnormal{smooth}_{\textnormal{L}_1}(s_t - s_c) + \textnormal{smooth}_{\textnormal{L}_1}(e_t -e_c)
%\end{equation}

\subsection {Implementation Details}
%We use TensorFlow to implement our model and utilize Tesla K80 GPU to accelerate the procedure of training and testing.
Our model is implemented using TensorFlow and deployed on the Tesla K80 GPU.
For a multi-scale deterministic BiPN model, both forward encoder and reverse encoder consist of $3$ convolutional layers while the decoder contains $4$ up-convolutional layers. The sizes of convolution kernels are diverse at different depths and scales. For multi-scale non-deterministic BiPN model, the basic architecture is almost the same with the deterministic one but adding an additional 100-dimensional noise vector as input. The noise is first passed through a fully-connected layer, reshaped to a feature map, and then concatenated to the intermediate feature map of the forward and reverse encoders. All input frames are resized to $64\times 64$ in training while kept the original sizes during testing.
%Our code can be found in https://github.com/papersubmission123/vcip2017.git.

\begin{figure}[t]
  \centering
  %\fbox{\rule[-.5cm]{0cm}{2cm} \rule[-.5cm]{8cm}{0cm}}
  \includegraphics[height=1.9cm]{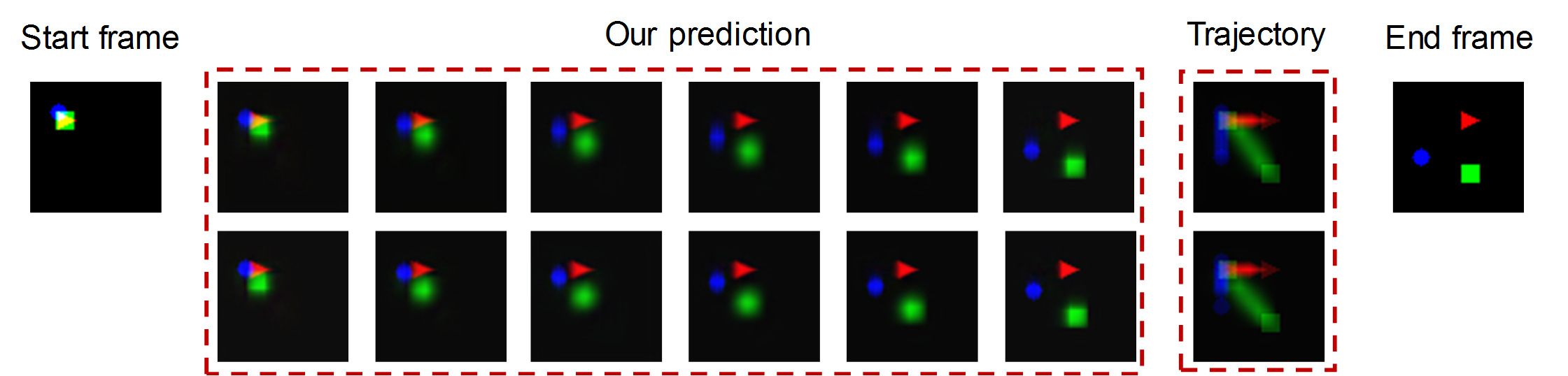}
  \caption{Two different sets of possible in-between frames are predicted by sampling different noise. Differences can be seen from two trajectory images.}\label{fig:fig-example-mul2d}
\end{figure}

\begin{figure*}[t]
  \centering
  %\fbox{\rule[-.5cm]{0cm}{4cm} \rule[-.5cm]{18cm}{0cm}}
  \includegraphics[height=2.5cm]{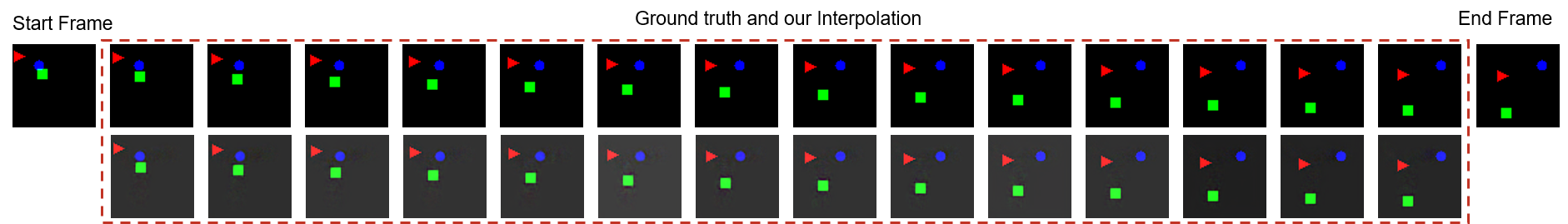}
  \caption{2D Shape results of interpolating $14$ frames. All objects of our interpolation in second row can move to the right positions as well as keep their original appearance unchanged compared to the ground truth in the first row.}\label{fig:fig-example-2d}
\end{figure*}
\begin{figure*}[t]
  \centering
  %\fbox{\rule[-.5cm]{0cm}{4cm} \rule[-.5cm]{18cm}{0cm}}
  \includegraphics[height=2.5cm]{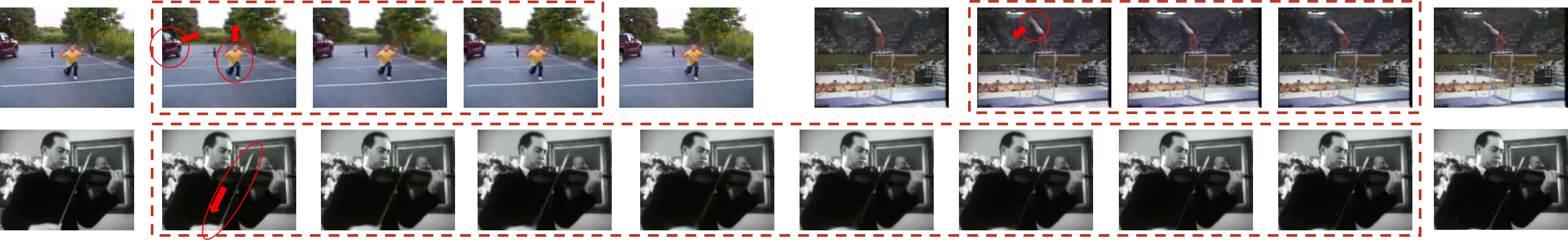}
  \caption{UCF101 results. First row: $2$ examples of interpolating $3$ frames. Notice the movements of the car, the weightlifter (ex$1$.) and the feet of the gymnast (ex$2$.). Second row: $1$ example of interpolating 8 frames that details how the violinist plays the violin. The red arrows indicate the moving directions.}\label{fig:fig-example-ucf}
  %\caption{UCF101 results. First row: $2$ examples of interpolating $3$ frames. Notice the movements of the car, the weightlifter (ex$1$.) and the feet of the gymnast (ex$2$.). Second row: $1$ example of interpolating 8 frames. We can see the surfer moves down and the waves roll.}\label{fig:fig-example-ucf}
\end{figure*}

\section{Experiments}
In this section, we evaluate our model on two datasets. We first conduct experiments using the dataset of synthetic 2D shapes to perform qualitative analysis. We then experiment our model for natural high-resolution videos using the UCF101 dataset and compare the results to recent outstanding methods.

\subsection{Moving 2D Shapes}
%Moving 2D shapes is a synthetic dataset provided by \cite{xue2016visual} for probabilistic future frame synthesis.
Moving 2D Shapes is a synthetic dataset containing three types of objects moving inside $64\times 64$ frames, where circles, squares and triangles move in random direction (vertically, horizontally or diagonally) with random velocity.  Fig. \ref{fig:fig-example-2d} shows our qualitative results with $14$ frames generated from only $2$ input frames. We can see that our interpolation results are quite close to the ground truth. In addition,  as can be seen in Fig. \ref{fig:fig-example-mul2d}, adding different noise samples to the non-deterministic model can produce multiple plausible motions as expected.

Experiments on this toy dataset demonstrate that our BiPN model can predict long-term in-between frames, capture the trajectories of different objects accurately and has the ability to produce multiple possible intermediate procedures.

\subsection{UCF101 Dataset}
The UCF101 dataset \cite{soomro2012ucf101} contains $13,320$ videos belonging to $101$ action categories and each video has a resolution of $240\times 320$. We sample $200,000$ video sequences from the training set to train BiPN and use the same test set with \cite{mathieu2015deep} \cite{liu2017video} as benchmarks. Both Peak Signal to Noise Ratio (PSNR) and Structural Similarity Index Measure (SSIM) \cite{wang2004image} are used to assess the image quality of the predicted frames, where higher PSNR and SSIM are better.

\begin{figure}[t]
  \centering
  %\fbox{\rule[-.5cm]{0cm}{3cm} \rule[-.5cm]{8cm}{0cm}}
  \includegraphics[height=2.5cm]{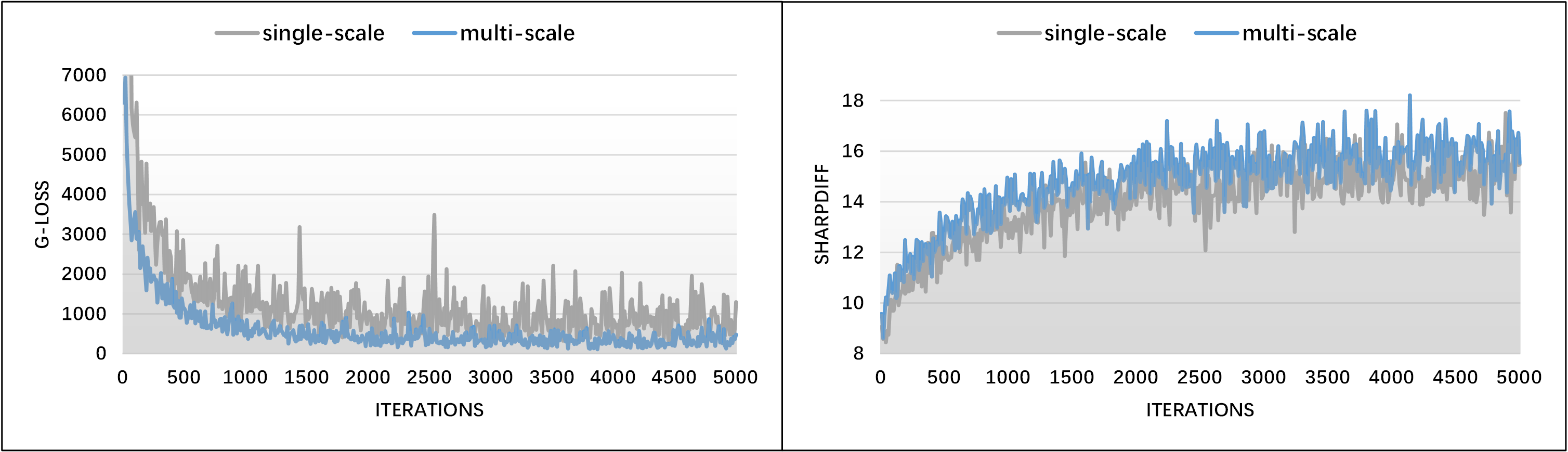}
  \caption{Generator loss and SharpDiff \cite{mathieu2015deep} (a measure for sharpness, higher is better) comparison of single-scale and multi-scale BiPN on UCF101 dataset.}\label{fig:fig-curve-ucf}
\end{figure}

Fig. \ref{fig:fig-curve-ucf} shows the change of generator loss and SharpDiff of the single-scale and multi-scale BiPN with different training iterations. We can see that the multi-scale BiPN obtains lower loss and higher SharpDiff than the single one. We further compare our approach against several outstanding methods, including the optical flow technique EpicFlow \cite{revaud2015epicflow} and DVF \cite{liu2017video}. To be comparable, we evaluate BiPN on single-frame generation as the published results of EpicFlow and DVF are obtained by only evaluating one in-between frame. The results are reported in Table \ref{tab:ucf}, which shows that multi-scale BiPN brings moderate improvement to the single-scale one. Also, our model can achieve better PSNR than EpicFlow and DVF as well as competitive performance in terms of SSIM.

Speculating the long-term procedures of events and further synthesizing frames in natural videos are rather difficult. Our BiPN can generate more than $8$ decent frames as shown in Fig. \ref{fig:fig-example-ucf}. More results can be found in our supplementary material.

%\begin{figure}[htb]
%  \begin{minipage}[b]{0.25\textwidth}
%   % \centering
%    \fbox{\rule[-.5cm]{0cm}{2cm} \rule[-.5cm]{4cm}{0cm}}
%    \caption{This is a Figure by a Table}
%    \label{fig:by:table}
%  \end{minipage}%
%  \begin{minipage}[b]{0.25\textwidth}
%  \caption{PSNR and SSIM comparison on UCF101.}
%  \label{tab:ucf}
%  \begin{tabular}{cccc}
%    \toprule
%    Model &PSNR & SSIM \\
%    \midrule
%    EpicFlow \cite{revaud2015epicflow} & 30.2 & 0.93 \\
%    DVF \cite{liu2017video} & 30.9 & \bftab 0.94 \\
%    Single-scale BiPN (ours) &  31.2 & 0.92 \\
%    Multi-scale BiPN (ours)&  \bftab 31.9 & \bftab 0.94 \\
%    \bottomrule
%      \end{tabular}
%  \end{minipage}
%\end{figure}

\section{Conclusion}
This paper introduces a bidirectional predictive network (BiPN) for long-term video interpolation. BiPN is built as an encoder-decoder model that can predict the intermediate frames in two opposite direction. A multi-scale version of BiPN is also developed to capture both small and large motions. To enable the model to imagine multiple possible motions, we feed the BiPN with an extra noise vector input as exploratory experiment. We demonstrate the advantages of BiPN on Moving 2D Shapes and UCF101 dataset and report competitive results to the recent approaches.

\begin{table}[t]\scriptsize
  \centering
  \caption{PSNR and SSIM comparison on UCF101.}
  \label{tab:ucf}
  \begin{tabular}{cccc}
    \toprule
    Model &PSNR & SSIM \\
    \midrule
    EpicFlow \cite{revaud2015epicflow} & 30.2 & 0.93 \\
    DVF \cite{liu2017video} & 30.9 & \bftab 0.94 \\
    Single-scale BiPN (ours) &  30.6 & 0.92 \\
    Multi-scale BiPN (ours)&  \bftab 31.4 & \bftab 0.94 \\
    \bottomrule
  \end{tabular}
\end{table}

\bibliographystyle{IEEEtran}

\bibliography{jz_ref}
%\bibliography{IEEEabrv,IEEEexample, ref}

\end{document}